\documentclass[letterpaper]{article}
\usepackage[preprint]{match4annotate-layout}
\usepackage[hyphens]{url}
\usepackage{graphicx}
\usepackage{natbib}
\usepackage{caption}
\usepackage{amsmath,amsfonts,amssymb}
\usepackage{mathtools}
\usepackage{booktabs}
\usepackage{multirow}
\usepackage{enumitem}
\usepackage[pdfencoding=auto,psdextra,hidelinks]{hyperref}
\hypersetup{
pdftitle={Match4Annotate: Cross-Video Annotation Transfer in Ultrasound via Implicit Feature Flow-Guided Matching},
pdfauthor={Zhuorui Zhang; Roger Pallares-Lopez; Praneeth Namburi; Brian W. Anthony}
}
\AtBeginDocument{
  \setcounter{topnumber}{3}
  \setcounter{totalnumber}{5}
  \setcounter{dbltopnumber}{3}

  \setlength{\floatsep}{5pt plus 2pt minus 2pt}
  \setlength{\textfloatsep}{6pt plus 2pt minus 2pt}
  \setlength{\intextsep}{6pt plus 2pt minus 2pt}
  \setlength{\dblfloatsep}{5pt plus 2pt minus 2pt}
  \setlength{\dbltextfloatsep}{6pt plus 2pt minus 2pt}
}
\newsavebox{\rightcoltables}
\newsavebox{\pointsbox}
\newsavebox{\masksbox}
\DeclareMathOperator*{\argmax}{argmax}
\newcommand{\eg}{e.g.}
\newcommand{\ie}{i.e.}

\title{Match4Annotate: Cross-Video Annotation Transfer in Ultrasound\\via Implicit Feature Flow-Guided Matching}
\author{
    Zhuorui Zhang\textsuperscript{1}\equalcontrib, Roger Pallar\`es-L\'opez\textsuperscript{1}\equalcontrib, Praneeth Namburi\textsuperscript{2,3}, and Brian W. Anthony\textsuperscript{1,2,3}
}
\affiliations{
\textsuperscript{1} Department of Mechanical Engineering, Massachusetts Institute of Technology, Cambridge, USA\\
\textsuperscript{2} Institute for Medical Engineering and Science, Massachusetts Institute of Technology, Cambridge, USA\\
\textsuperscript{3} MIT.nano Immersion Lab, Massachusetts Institute of Technology, Cambridge, USA\\
\href{mailto:banthony@mit.edu}{banthony@mit.edu}
}
\begin{document}
\maketitle
\begin{figure*}[!t]
  \centering
  \includegraphics[width=0.99\linewidth]{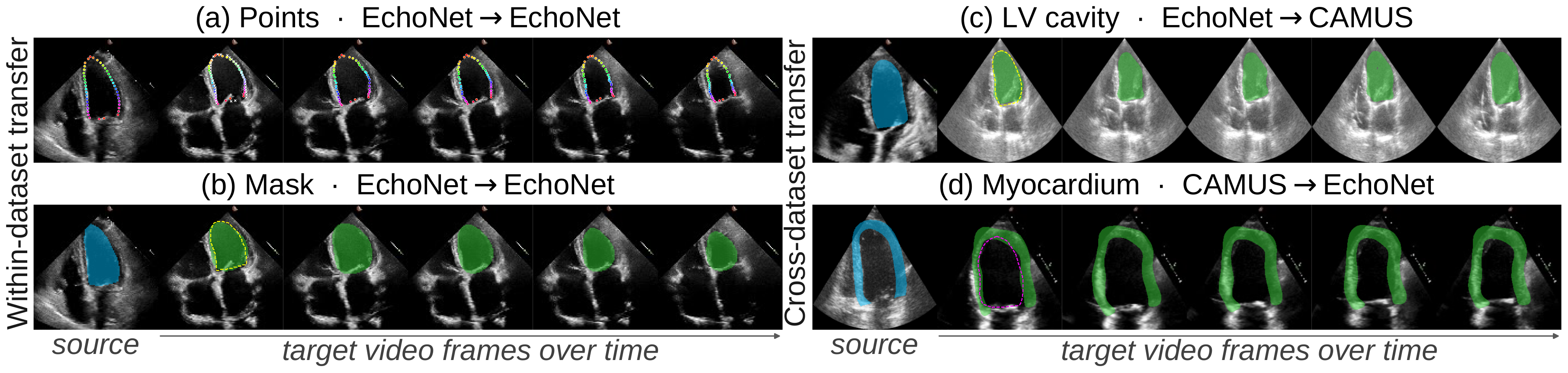}
  \caption{\textbf{Overview of Match4Annotate: annotation transfer across videos and datasets, shown over time.} A single source annotation (left) is transferred to a \emph{different} target video and shown \emph{across time} as a temporal strip of target frames. \textbf{Within-dataset} (source and target both EchoNet): \textbf{(a)}~Boundary-point tracking (colors encode correspondence identity across frames) and \textbf{(b)}~LV-cavity mask transfer; ground truth (white $\times$ and yellow dashed contour) is shown on the labeled target frame. \textbf{Cross-dataset}: \textbf{(c)}~The LV cavity, annotated in both datasets, is transferred EchoNet$\rightarrow$CAMUS (ground-truth contour in yellow), and \textbf{(d)}~The myocardium, annotated in CAMUS but \emph{not} EchoNet, is borrowed from CAMUS and transferred onto EchoNet; with no target ground truth, EchoNet's own LV-cavity contour (magenta) is shown for anatomical context. Predicted masks in green, source annotations in cyan.}
  \label{fig:1}
\end{figure*}
\begin{abstract}
Acquiring per-frame annotations for ultrasound videos is costly and requires substantial clinical expertise, limiting the development of learning-based analysis methods. We study \emph{cross-video annotation transfer}: propagating user-specified annotations from a labeled ultrasound video to an independently acquired target video with no target-side labels or manual initialization. Video trackers and segmentation propagators rely on temporal continuity and require a prompt in every new sequence, whereas cross-image feature matching and one-shot segmentation estimate correspondences independently, without enforcing spatially coherent deformations or providing unified support for point and mask annotations. We present Match4Annotate, a framework that addresses this setting at test time through three stages. A \emph{spatiotemporal implicit feature representation} lifts frozen vision foundation-model features into a continuous field over space and time, enabling queries beyond the native resolution of the backbone. A \emph{continuous implicit feature flow} then aligns the source and target fields under a smooth-deformation prior, estimating correspondence in feature space rather than relying on intensity consistency, which is often violated in ultrasound by speckle and acquisition-dependent appearance. Finally, \emph{flow-guided annotation transfer} uses the estimated flow as a spatial prior over feature similarity. This formulation unifies sparse point and dense mask transfer and includes unconstrained feature matching and direct flow warping as limiting cases. On four clinical ultrasound datasets spanning echocardiography and musculoskeletal imaging, Match4Annotate achieves state-of-the-art annotation transfer, outperforming dense feature-matching baselines across PCK thresholds and one-shot segmentation methods in Dice score. It also demonstrates bidirectional transfer of left-ventricular annotations across echocardiography datasets. It requires no task-specific training and adapts to each video in minutes on a single consumer GPU.
\end{abstract}
\section{Introduction}
\label{sec:intro}
The scarcity of per-frame annotations, such as segmentation masks or anatomical point trajectories, remains a major obstacle to scaling learning-based analysis of ultrasound videos. Expert medical annotation can cost approximately \$330--610 per hour of radiologist time~\cite{ucdavis2024research}: annotating all 10,030 videos of EchoNet-Dynamic~\cite{ouyang2020video} at 32 frames per video would require roughly 1,900 expert hours, beyond the practical resources of most research groups and clinical institutions.
A natural strategy to reduce annotation effort is label propagation: given a small number of user annotations, automatically transfer them to additional frames. Zero-shot trackers and video object segmentation models such as SAM~2~\cite{ravi2024sam2}, CoTracker3~\cite{karaev2024cotracker3}, and XMem~\cite{cheng2022xmem} have made this practical within individual videos~\cite{tapnext,trackon2}. However, these methods rely on temporal continuity and require separate initialization for each video through points, masks, or memory prompts, making them unsuitable for annotation transfer across independently acquired ultrasound videos. We distinguish this manual initialization, which requires additional user input for every video, from automatic test-time optimization, which requires computation but no further annotation effort.
Classical correspondence pipelines and learned local-feature methods (\eg, SuperPoint~\cite{superpoint}, DISK~\cite{disk}, LightGlue~\cite{lightglue}, XFeat~\cite{xfeat}) match reliably but operate on detector-selected keypoints, making it impractical to transfer user-specified points or dense masks at arbitrary locations, especially in low-texture, low-contrast ultrasound where local appearance gives weak cues. Foundation-model correspondence methods~\cite{roma2024,dift,matcha} exploit semantically rich representations and do enable user-queryable point transfer across independently acquired images. Reliable annotation transfer, however, requires more than pairwise matching: it demands a dense, spatially coherent mapping that supports both points and masks and remains robust to the domain gap between ultrasound and natural-image pretraining.
Together, these gaps define a distinct problem setting: transferring user-specified annotations from a labeled ultrasound video to an independently acquired target video, without target-side labels or manual initialization, while supporting both sparse points and dense masks. We call this setting \emph{cross-video annotation transfer}. To our knowledge, prior ultrasound work has not jointly formulated and benchmarked target-initialization-free transfer of both user-specified points and masks across independently acquired videos.
\subsubsection{Our Approach.}
We introduce \textit{\textbf{Match4Annotate}}, a lightweight framework tailored to this setting. Its design is motivated by the properties of ultrasound that remain reliable across acquisitions. Fine-scale image appearance is strongly affected by speckle, probe orientation, gain, and operator-dependent acquisition, making image-intensity consistency unreliable for cross-video correspondence. In contrast, the underlying anatomical organization is more stable. We therefore establish correspondence in the feature space of self-supervised DINO models, whose representations retain useful anatomical structure despite being pretrained primarily on natural images. We then use implicit neural representations to model both the spatiotemporal feature field and the source--target flow continuously. Their smooth coordinate-based parameterization regularizes these fields, while SIREN activations provide stable spatial derivatives for optimizing the flow through feature alignment and smoothness constraints. Match4Annotate transfers annotations from labeled source videos to unlabeled target videos automatically, requiring no task-specific offline training or target-side user input, and adapts to each video in minutes on a single consumer GPU. Our contributions are:
\begin{enumerate}[leftmargin=*,nosep]
\item \textbf{Cross-video annotation transfer for ultrasound.}
We formulate this setting and instantiate it in Match4Annotate, which transfers both point and mask annotations through a unified test-time framework. We evaluate on four clinical ultrasound datasets spanning echocardiography and musculoskeletal imaging and, to our knowledge, provide the first benchmark for target- initialization-free transfer of both annotation types across independently acquired videos. We further demonstrate bidirectional transfer between EchoNet-Dynamic \cite{ouyang2020video} and CAMUS-TED \cite{painchaud2022temporal}, including propagation of myocardial masks to EchoNet, where that structure is not annotated.
\item \textbf{Spatiotemporal implicit feature representation.} For each video, we fit a SIREN~\cite{sitzmann2020siren} to frozen DINOv3~\cite{dinov3} features jointly over space and time. Unlike feature upsamplers such as FeatUp~\cite{fu2023featup} and AnyUp~\cite{wimmer2025anyup}, which recover finer spatial detail for downstream dense prediction, our field serves as a continuous substrate for correspondence: it can be queried and differentiated with respect to position at arbitrary sub-patch coordinates. Under the same spatial prior, it outperforms both upsamplers (Sec.~\ref{sec:ablation}).
\item \textbf{Continuous implicit feature flow.}
A second implicit network predicts a dense flow field between source and target frames. We optimize this deformation by aligning the two continuous feature fields, so correspondence is driven by semantic feature consistency rather than image-intensity agreement, which is often unreliable in ultrasound because of speckle and acquisition-dependent appearance. The flow formulation is not tied to our feature representation: when applied to FeatUp and AnyUp, it improves Dice by $13.1$ and $7.8$ points, respectively (Sec.~\ref{sec:ablation}), demonstrating that it contributes substantially across all tested feature representations.
\end{enumerate}
\emph{Flow-guided annotation transfer} combines the two components into the final correspondence mechanism: the estimated flow defines a spatial prior that reweights feature similarity, rather than directly warping the annotations. The same formulation transfers both sparse points and dense masks, while unconstrained feature matching is recovered when the prior becomes uninformative.
\section{Related Work}
\label{sec:related}
\noindent\textbf{Implicit Neural Representations.}
Implicit neural representations map coordinates to signal values, yielding continuous, resolution-independent fields. SIREN~\cite{sitzmann2020siren} uses sinusoidal activations to capture fine detail while retaining smooth analytical derivatives. INRs are widely used for 3D neural rendering~\cite{mildenhall2020nerf,muller2022instant} and video representation, where they typically model RGB appearance over space and time~\cite{chen2022videoinr}. Coordinate-based networks have also been applied to image registration~\cite{wolterink2022idir} and dynamic scenes~\cite{pumarola2021d}. When correspondence is optimized through intensity or radiance agreement, however, it relies on appearance consistency that is often unreliable in ultrasound. Speckle reflects wave interference rather than tissue appearance directly, while echo amplitude varies with probe orientation, attenuation, and acquisition settings. We instead optimize a continuous flow field through semantic feature alignment, allowing correspondence to follow anatomical structure despite substantial appearance variation across videos and datasets.
\noindent\textbf{Feature Upsampling and Foundation Models.}
Vision foundation models such as DINO~\cite{caron2021emerging} and DINOv2~\cite{oquab2023dinov2} provide semantically rich representations, but their patch-based outputs limit spatial precision. Feature-upsampling methods, including FeatUp~\cite{fu2023featup}, LoftUp~\cite{huang2025loftup}, and AnyUp~\cite{wimmer2025anyup}, recover finer detail but operate primarily on individual images. DINOv3~\cite{dinov3} improves dense feature quality, but neither backbones nor upsamplers enforce temporal consistency, which fitting a SIREN over spatio-temporal coordinates supplies.
\noindent\textbf{Video Point and Mask Tracking.}
Optical-flow methods such as SEA-RAFT~\cite{wang2024searaft}, WAFT~\cite{wang2025waft}, and GMFlow++~\cite{xu2023unimatch} estimate dense motion between image pairs, while long-term point trackers including TAPNext~\cite{tapnext}, CoTracker3~\cite{karaev2024cotracker3}, Track-On2~\cite{trackon2}, and DINO Tracker~\cite{tumanyan2024dinotracker} extend correspondence across video frames. Similarly, STCN~\cite{cheng2021rethinking}, XMem~\cite{cheng2022xmem}, and SAM~2~\cite{ravi2024sam2} propagate masks through temporally connected sequences. These
methods rely on temporal continuity and are not designed to transfer annotations across independently acquired videos without manual target initialization.
\noindent\textbf{Cross-Image Annotation Transfer.}
Foundation-model correspondence methods such as RoMa~\cite{roma2024}, DIFT~\cite{dift}, and MATCHA~\cite{matcha} support matching across independently acquired images, but do not explicitly enforce spatial and temporal coherence throughout a target video. In-context segmentation methods such as UniverSeg~\cite{univseg} and Matcher~\cite{matcher} transfer masks from labeled examples, but do not recover arbitrary point correspondences.
\noindent\textbf{Test-Time Optimization.}
Test-time methods adapt models or auxiliary parameters using unlabeled target data~\cite{sun2020test,wang2021tent}. Match4Annotate keeps the foundation model frozen and fits lightweight implicit fields per video, requiring neither offline training nor target-video annotations.
\section{Method}
\label{sec:method}
\subsection{Overview}
\begin{figure*}[!t]
  \centering
  \includegraphics[width=0.99\linewidth]{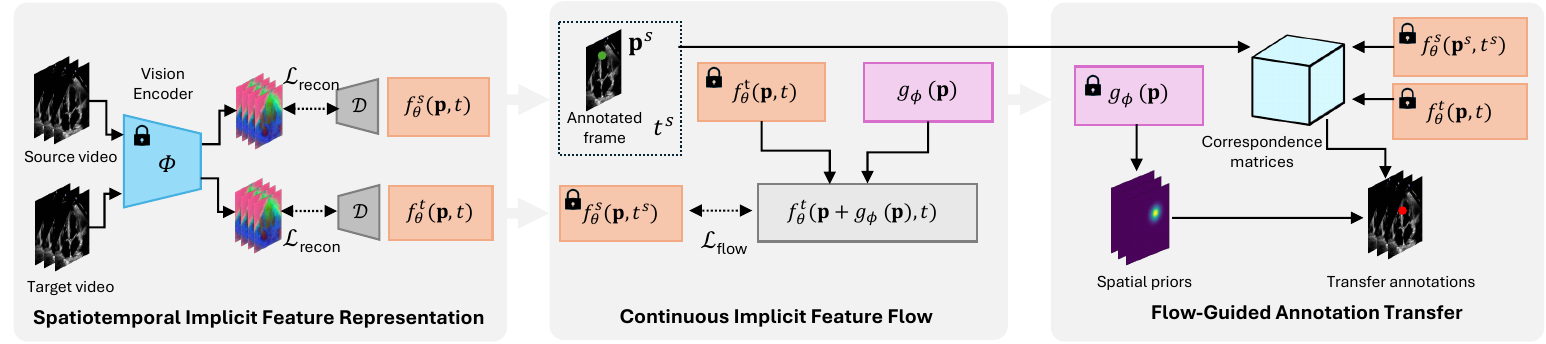}
\caption{\textbf{Match4Annotate workflow.} \textbf{Spatiotemporal implicit feature representation} fits a per-video implicit field to frozen DINOv3 features, yielding a continuous representation that can be queried at arbitrary spatial and temporal coordinates. \textbf{Continuous implicit feature flow} then estimates a smooth spatial displacement between a source and a target frame from these representations. \textbf{Flow-guided annotation transfer} converts that displacement into a spatial prior on feature-similarity matching, transferring annotated points and masks onto the target. Padlocks mark modules frozen at that stage.}
  \label{fig:method}
\end{figure*}

Given a video $\mathcal{V} = \{I_1, \ldots, I_T\}$ and annotations on a source frame, either individual points of interest (POI) or a binary segmentation mask, our goal is to transfer these annotations to target frames across different videos of the same anatomy. Throughout, $\mathbf{p} \in \Omega = [1,W] \times [1,H]$ denotes a continuous spatial coordinate. Match4Annotate proceeds in three stages (Figure~\ref{fig:method}). \emph{Spatiotemporal implicit feature representation} (Sec.~\ref{sec:feature}) fits a SIREN to frozen vision foundation model (VFM) features, giving one continuous feature field per video that can be queried at arbitrary resolution. \emph{Continuous implicit feature flow} (Sec.~\ref{sec:flow}) optimizes a second, lightweight SIREN $g_\phi$ that aligns the source and target fields, yielding a smooth flow field between them. \emph{Flow-guided annotation transfer} (Sec.~\ref{sec:transfer}) converts that flow into a spatial prior on feature-similarity matching; masks are transferred by applying the same procedure to interior points (Sec.~\ref{sec:mask}). We summarize all implementation details discussed in this section in the Supplementary Materials.
\subsection{Feature Extraction}
We extract dense features from each frame $I_t$ using DINOv3-ViT-S/16~\cite{dinov3} as our frozen VFM $\Phi$, producing a grid of patch tokens:
\begin{equation}
F_t = \Phi(I_t) \in \mathbb{R}^{H' \times W' \times D}
\label{eq:feat}
\end{equation}
where $H' = W' = S/p$ is determined by the input size $S$ and patch size $p$, and $D$ is the feature dimension. All features are $\ell_2$-normalized. The VFM is used only for feature extraction and is not fine-tuned.
\subsection{Spatiotemporal Implicit Feature Representation}
\label{sec:feature}
\subsubsection{Problem Formulation.}
Given video features $\mathcal{F} = \{F_1, \ldots, F_T\}$, we learn a continuous function $f_\theta : \Omega \times [1,T] \rightarrow \mathbb{R}^D$ that maps any spatiotemporal coordinate to a feature vector, where $\theta$ denotes the learnable parameters:
\begin{equation}
f_\theta(\mathbf{p}, t) \in \mathbb{R}^D, \quad (\mathbf{p}, t) \in \Omega \times [1, T]
\label{eq:implicit}
\end{equation}
enabling queries at arbitrary resolution and dense correspondence at sub-patch granularity. One field is fit per video: following Figure~\ref{fig:method}, we write $f^s_\theta$ and $f^t_\theta$ for the fields of the source and target videos, and $t^s$ for the annotated source frame.
\subsubsection{Network Architecture.}
We adopt SIREN~\cite{sitzmann2020siren} with periodic activations $\rho(a) = \sin(\omega a)$, where $\omega$ is the frequency hyperparameter. Sinusoidal activations give frequency-controllable representations that mitigate interpolation artifacts at arbitrary query resolutions, yield well-behaved higher-order derivatives for stable gradient-based flow estimation, and provide an inductive bias for cyclic anatomical motion. We quantitatively compare SIREN against ReLU, ReLU+PE, and Instant-NGP in the Supplementary Materials. The network has $L$ fully-connected layers with $H_\text{dim}$ hidden dimensions. Different from FeatUp~\cite{fu2023featup}, we use coordinates only $(\mathbf{p},t)$ without intensity $I_t(\mathbf{p})$ to avoid spurious correlations from ultrasound speckle noise, prioritizing anatomy-level semantic consistency over texture fidelity.
\subsubsection{Training Objective.}
We train $f_\theta$ to reproduce the VFM features at high resolution via a learned downsampler. We evaluate $f_\theta$ on a dense grid $\Omega_\text{HR}$, then a convolutional downsampler $\mathcal{D}$ maps the output back to the VFM resolution:
\begin{equation}
\mathcal{L}_\text{recon} = \frac{1}{N} \sum_{i=1}^{N} \big\| \mathcal{D}(f_\theta(\mathbf{p}_i, t_i)) - F_{t_i}(\mathbf{p}_i) \big\|_2^2
\label{eq:recon}
\end{equation}
where $\mathcal{D}$ is a learnable depthwise convolution with non-negative, normalized weights. We choose $\Omega_\text{HR}$ with size $H \times W$ for downstream pixel-level matching purposes and optimize with Adam~\cite{kingma2014adam}. After training, $f_\theta$ can be queried at any continuous coordinate $(\mathbf{p},t)$.
\subsection{Continuous Implicit Feature Flow}
\label{sec:flow}
We estimate a dense, spatially coherent deformation between the source and target videos entirely in feature space. A lightweight SIREN $g_\phi:\Omega\rightarrow\mathbb{R}^2$ predicts a displacement at each spatial coordinate,
\begin{equation}
\Delta\mathbf{p} = g_\phi(\mathbf{p}),
\label{eq:disp}
\end{equation}
so that $\mathbf{p} \mapsto \mathbf{p} + g_\phi(\mathbf{p})$ maps source coordinates to their target counterparts. Because $f^t_\theta$ is continuous, it can be queried directly at the displaced coordinates. We optimize $g_\phi$ to align the displaced target field with the source field over a dense query set $\Omega_\text{HR}$, while keeping the flow smooth:
\begin{equation}
\begin{split}
\mathcal{L}_\text{flow} = {}& \frac{1}{|\Omega_\text{HR}|} \sum_{\mathbf{p} \in \Omega_\text{HR}} \big\| f^t_\theta(\mathbf{p} {+} g_\phi(\mathbf{p}),\, t) \\
&\quad - f^s_\theta(\mathbf{p},\, t^s) \big\|_2^2 + \lambda_1 \mathcal{L}_\text{TV} + \lambda_2 \mathcal{L}_\text{mag}
\end{split}
\label{eq:flow}
\end{equation}
where $\mathcal{L}_\text{TV}$ is total-variation regularization~\cite{rudin1992nonlinear} encouraging spatial smoothness of the flow field, and $\mathcal{L}_\text{mag} = \frac{1}{|\Omega_\text{HR}|}\sum_{\mathbf{p}} \|g_\phi(\mathbf{p})\|_1$ penalizes displacement magnitude, avoiding unnecessary deformation on flat regions.
\subsection{Flow-Guided Annotation Transfer}
\label{sec:transfer}
Given source annotations $\{\mathbf{p}^s_i\}_{i=1}^K$ on frame $t^s$, the estimated flow guides each of them in the target,
\begin{equation}
\hat{\mathbf{p}} = \mathbf{p}^s + g_\phi(\mathbf{p}^s).
\label{eq:phat}
\end{equation}
Rather than accept $\hat{\mathbf{p}}$ directly, we treat it as a \emph{spatial prior} that reweights feature similarity. Let $\kappa:\mathbb{R}^2\rightarrow(0,1]$ be a prior kernel that is radially non-increasing with $\kappa(\mathbf{0}) = 1$. The transferred point is
\begin{equation}
\begin{split}
\mathbf{p}^* = \argmax_{\mathbf{p} \in \Omega_\text{HR}} \; & \cos\!\big(f^s_\theta(\mathbf{p}^s, t^s),\, f^t_\theta(\mathbf{p}, t)\big) \\
& \cdot\, \kappa(\mathbf{p} - \hat{\mathbf{p}})
\end{split}
\label{eq:match}
\end{equation}
where $\cos(\cdot,\cdot)$ is cosine similarity and $\Omega_\text{HR}$ densely samples the target frame. The kernel encodes how much spatial evidence is trusted relative to appearance: $\kappa \equiv 1$ recovers unconstrained nearest-neighbour matching, while a kernel concentrated at $\hat{\mathbf{p}}$ recovers pure flow warping. Between these extremes it combines the large-displacement capability of the flow with the discriminative power of feature matching. We instantiate $\kappa$ as an isotropic Gaussian whose bandwidth adapts to the annotation canvas size. Sec.~\ref{sec:ablation} ablates the unconstrained case $\kappa \equiv 1$ and a mis-guided prior.
\subsection{Mask Transfer via Interior Point Method}
\label{sec:mask}
Rather than transfering only contour points and reconstructing masks from noisy boundary predictions, we use an interior-point method that densely samples the annotated region.
\subsubsection{Interior Point Generation.}
Given a source mask $M^s$, we extract interior points using the Euclidean distance transform:
\begin{equation}
\mathcal{I}^s = \big\{ \mathbf{p} \in \Omega \;\big|\; \text{EDT}(\mathbf{p};\, M^s) \geq d_\text{min} \big\}
\label{eq:edt}
\end{equation}
where $\text{EDT}(\cdot)$ is the distance to the nearest background pixel and $d_\text{min}$ controls the margin from the boundary, excluding ambiguous boundary pixels.
\subsubsection{Point Propagation and Reconstruction.}
All interior points are propagated to the target by flow-guided transfer (Sec.~\ref{sec:transfer}), producing $\mathcal{I}^t$. These are converted to a mask by kernel density estimation: we place unit impulses at the propagated points and convolve with a smoothing kernel $k$,
\begin{equation}
P(\mathbf{p}) = \frac{1}{Z} \sum_{\mathbf{p}^*_j \in \mathcal{I}^t} k\big(\mathbf{p} - \mathbf{p}^*_j\big),
\label{eq:kde}
\end{equation}
where $Z$ normalizes $P$ to $[0,1]$, followed by thresholding $\hat{M}^t(\mathbf{p}) = \mathbf{1}[P(\mathbf{p}) \geq \tau]$. As for the prior kernel, we instantiate $k$ as an isotropic Gaussian, whose bandwidth $\sigma_\text{KDE}$ is tuned jointly with $\tau$ by grid search (additional details in the Supplementary Materials). Dense interior sampling provides redundancy: the kernel smooths individual point errors, so the mask degrades gracefully when a subset of points are mismatched.
\section{Experiments}
\label{sec:experiments}
\begin{figure*}[t]
\centering
\includegraphics[width=0.99\linewidth]{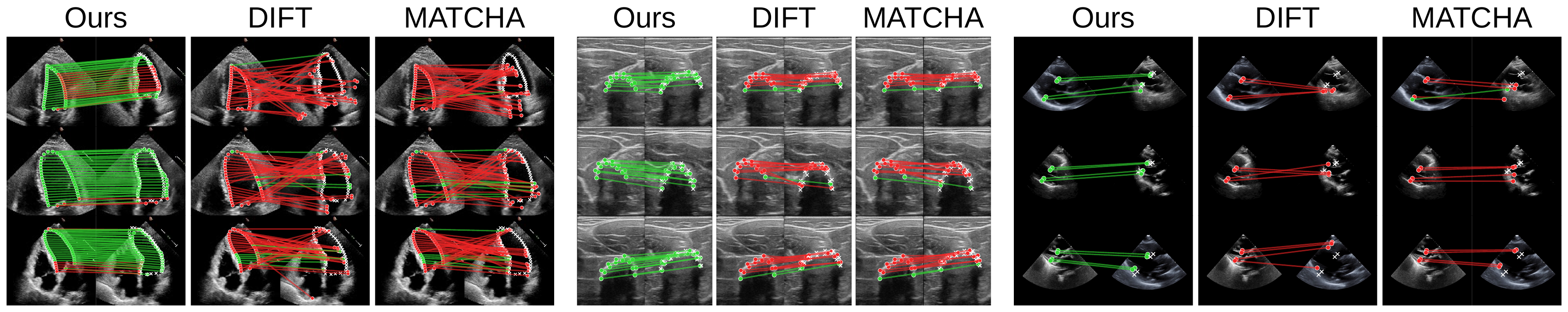}
\caption{\textbf{Cross-video point transfer} on EchoNet (left), MSK-Bone (middle), and EchoNet-LVH (right) comparing with DIFT and MATCHA. Each cell shows a source frame with annotated points and a target frame with transferred predictions; green lines indicate correct correspondences (within the PCK radius) and red lines indicate incorrect ones.}
\label{fig:poi_comparison}
\end{figure*}
\subsection{Experimental Setup}
\label{sec:experimental_setup}
\subsubsection{Datasets.}
We evaluate on four ultrasound datasets spanning cardiac and musculoskeletal imaging. \emph{EchoNet-Dynamic}~\cite{ouyang2020video} provides apical four-chamber videos with ordered left-ventricular tracing points at end-diastole and end-systole; \emph{MSK-Bone}, a 20-subject subset from a public 36-subject upper-arm dataset~\cite{pallareslopez2026multimodal}, expert humerus masks and ordered boundary points; \emph{EchoNet-LVH}~\cite{duffy2022lvh}, parasternal long-axis videos with IVS, LVID, and LVPW endpoints; and \emph{CAMUS-TED}~\cite{painchaud2022temporal}, full-cycle left-ventricular masks.
\subsubsection{Evaluation Protocol.}
We sample 200 source--target video pairs from different subjects for
EchoNet-Dynamic, EchoNet-LVH, and CAMUS-TED, and evaluate 380 unique subject pairs for MSK-Bone. Because EchoNet-Dynamic and
CAMUS-TED share left-ventricular landmarks, we also assess
cross-dataset transfer on 200 pairs drawn between them.
\subsubsection{Implementation Details.}
Both implicit networks are optimized per video at test time with Adam~\cite{kingma2014adam}, with the vision foundation model frozen throughout. Complete hyperparameters are provided in the Supplementary Materials, including network width and depth, the frequency $\omega$, the loss weights $\lambda_1$ and $\lambda_2$, the interior margin $d_\text{min}$, and the kernel bandwidths.
\subsubsection{Evaluation Metrics}
We evaluate point and mask transfer using complementary metrics. For point transfer, we report the Percentage of Correct Keypoints (PCK) at thresholds of $\{4,8,16\}$ pixels after resizing all frames and coordinates to a $256{\times}256$ canvas (\ie, PCK@$ \{.016,.031,.063\}$ of the image width). For mask transfer, we report the Dice coefficient between the thresholded KDE map (Sec.~\ref{sec:mask}) and the ground-truth mask.
\subsubsection{Baselines.}
We compare against specialized baselines for point matching and mask
segmentation. For point transfer, we evaluate the dense
correspondence methods RoMa~\cite{roma2024},
DIFT~\cite{dift}, and MATCHA~\cite{matcha}. For mask transfer, we
compare against Matcher~\cite{matcher} using SemanticSAM and SAM
backbones, and UniverSeg~\cite{univseg} in both 1-shot and
multi-shot settings; multi-shot results are marked with an asterisk
because they use additional labeled support examples.
\subsection{Point \& Mask Annotation Transfer}
\label{sec:inter}
\begin{figure*}[t]
\centering
\includegraphics[width=0.99\linewidth]{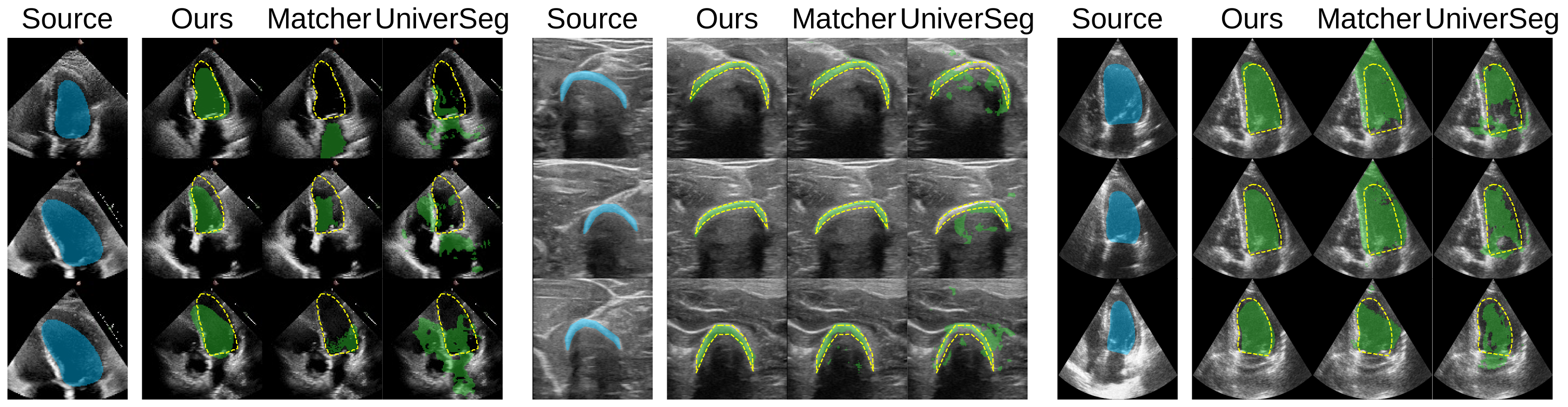}
\caption{\textbf{Cross-video mask transfer} on EchoNet (left), MSK-Bone (middle), and CAMUS-TED (right) comparing with Matcher (SemSAM) and UniverSeg (1-shot). Each cell shows the target frame with the transferred mask (green) and ground-truth contour (yellow dashed); the leftmost column of each panel shows the source reference mask (cyan).}
\label{fig:mask_comparison}
\end{figure*}

Tables~\ref{tab:inter_pck} and~\ref{tab:inter_dice} report the main cross-video results, with qualitative comparisons in
Figures~\ref{fig:poi_comparison} and~\ref{fig:mask_comparison}. For point correspondence, Match4Annotate achieves the best result in seven of the nine dataset--threshold combinations. It leads at all
thresholds on EchoNet, at PCK@$\{.031,.063\}$ on MSK-Bone, and at
PCK@$\{.016,.063\}$ on EchoNet-LVH; DIFT leads only at the strictest MSK-Bone threshold and MATCHA at PCK@$.031$ on EchoNet-LVH. The largest gains appear at the coarser thresholds, where Match4Annotate improves PCK@$.063$ from $41.1$ to $46.7$ on EchoNet and from $66.7$ to $74.3$ on MSK-Bone.
For mask transfer, Match4Annotate outperforms every one-shot segmentation baseline on all three datasets. From one annotated source frame it matches UniverSeg 5-shot on EchoNet, surpasses it on MSK-Bone, stays within $0.7$ Dice on CAMUS-TED, and trails 10-shot by only $2.4$--$4.4$ Dice. Per-dataset breakdowns and additional qualitative examples for all four datasets are provided in the Supplementary Materials.
\begin{table}[!t]
\centering
\caption{\textbf{Cross-video point transfer.} PCK@$\{.016,.031,.063\}\uparrow$ on EchoNet (200), MSK-Bone (380), and EchoNet-LVH (200 pairs). \textbf{Bold} best, \underline{underline} second.}
\label{tab:inter_pck}
{\small\setlength{\tabcolsep}{1pt}
\begin{tabular}{lccc}
\toprule
Method & EchoNet & MSK-Bone & EchoNet-LVH \\
\midrule
RoMa (in.)  & $5.2/15.8/39.2$ & $8.1/20.0/37.9$ & $4.5/33.4/64.2$ \\
RoMa (out.) & $\underline{5.5}/17.1/\underline{41.1}$ & $6.9/17.8/34.5$ & $3.8/29.5/59.8$ \\
DIFT           & $4.0/11.7/27.2$ & $\mathbf{16.3}/\underline{36.6}/62.4$ & $4.4/29.5/49.9$\\
MATCHA         & $5.7/\underline{17.8}/40.8$ & $13.5/36.4/\underline{66.7}$ & $3.9/\mathbf{39.5}/\underline{68.4}$ \\
\midrule
Ours & $\mathbf{6.1/18.4/46.7}$ & $\underline{15.5}/\mathbf{40.0/74.3}$ & $\mathbf{11.9}/\underline{34.7}/\mathbf{69.5}$ \\
\bottomrule
\end{tabular}}
\end{table}
\begin{table}[!t]
\centering
\caption{\textbf{Cross-video mask transfer.} Dice$\uparrow$ on EchoNet (200), MSK-Bone (380), and CAMUS-TED (200 pairs, all frames). $^\ast$multi-shot (not one-shot-comparable). \textbf{Bold} best, \underline{underline} second.}
\label{tab:inter_dice}
{\small\setlength{\tabcolsep}{1pt}
\begin{tabular}{lccc}
\toprule
Method & EchoNet & MSK-Bone & CAMUS-TED \\
\midrule
\multicolumn{4}{l}{\textit{Few-Shot Segmentation}} \\
\quad UniverSeg (1-shot)         & $47.1{\pm}16.2$ & $25.0{\pm}11.0$ & \underline{$57.4{\pm}13.9$} \\
\quad UniverSeg (3-shot)$^\ast$  & $73.0{\pm}14.0$ & $53.2{\pm}14.9$ & $76.3{\pm}10.5$ \\
\quad UniverSeg (5-shot)$^\ast$  & $76.3{\pm}12.5$ & $63.7{\pm}12.9$ & $79.2{\pm}9.8$\\
\quad UniverSeg (10-shot)$^\ast$ & $78.7{\pm}12.2$ & $73.4{\pm}8.9$  & $82.5{\pm}9.5$ \\
\midrule
\multicolumn{4}{l}{\textit{One-Shot Segmentation}} \\
\quad Matcher (SemSAM) & \underline{$53.8{\pm}21.7$} & \underline{$62.2{\pm}23.2$} & $55.6{\pm}14.9$ \\
\quad Matcher (SAM)         & $43.2{\pm}16.8$ & $26.0{\pm}29.4$ & $38.8{\pm}13.3$ \\
\midrule
Ours & $\mathbf{76.3{\pm}12.3}$ & $\mathbf{69.0{\pm}11.9}$ & $\mathbf{78.5{\pm}13.7}$ \\
\bottomrule
\end{tabular}}
\end{table}
\begin{table}[!t]
\centering
\caption{\textbf{Cross-dataset mask transfer.}
Dice$\uparrow$ for LV-cavity transfer between EchoNet-Dynamic and
CAMUS-TED.
EchoNet$\rightarrow$CAMUS evaluates all target frames, while
CAMUS$\rightarrow$EchoNet evaluates ED and ES frames.
$^\ast$Multi-shot results are not directly comparable.
Among one-shot methods, \textbf{bold} denotes best and
\underline{underline} second best.}
\label{tab:cross_dataset}
{\small\setlength{\tabcolsep}{1pt}
\begin{tabular}{lcc}
\toprule
Method
& EchoNet$\rightarrow$CAMUS
& CAMUS$\rightarrow$EchoNet \\
\midrule
\multicolumn{3}{l}{\textit{Few-Shot Segmentation}} \\
\quad UniverSeg (1-shot)
& $44.5{\pm}15.7$
& $42.6{\pm}15.6$ \\
\quad UniverSeg (3-shot)$^\ast$ & $68.7{\pm}13.1$ & $62.2{\pm}16.4$ \\
\quad UniverSeg (5-shot)$^\ast$
& $73.0{\pm}12.2$
& $66.5{\pm}16.3$ \\
\quad UniverSeg (10-shot)$^\ast$
& $76.5{\pm}11.1$
& $68.4{\pm}16.4$ \\
\midrule
\multicolumn{3}{l}{\textit{One-Shot Segmentation}} \\
\quad Matcher (SemSAM)
& \underline{$52.4{\pm}23.5$}
& \underline{$50.2{\pm}28.6$} \\
\quad Matcher (SAM)
& $39.5{\pm}18.4$
& $43.6{\pm}22.8$ \\
\midrule
Ours
& $\mathbf{71.9{\pm}16.0}$
& $\mathbf{66.8{\pm}17.0}$ \\
\bottomrule
\end{tabular}}
\end{table}
\subsection{Cross-Dataset Annotation Transfer}
\label{sec:cross_dataset}
\begin{figure}[t]
\centering
\includegraphics[width=\linewidth]{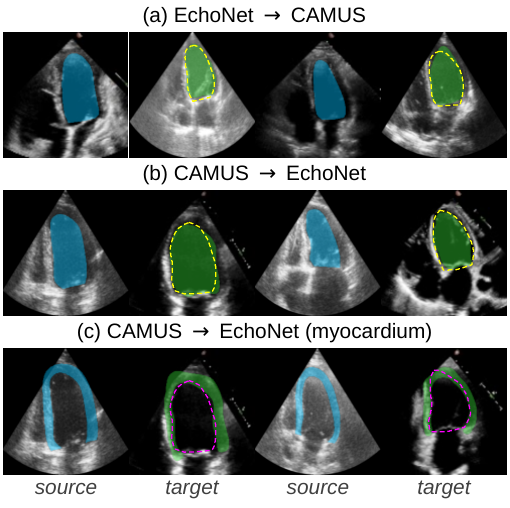}
\caption{\textbf{Cross-dataset annotation transfer between EchoNet and CAMUS-TED.} Each panel shows two source$\rightarrow$target examples; a source frame with its annotation (cyan) is followed by the target frame with the transferred mask (green). \textbf{(a)}~EchoNet$\rightarrow$CAMUS and \textbf{(b)}~CAMUS$\rightarrow$EchoNet transfer the shared left-ventricle cavity; the ground-truth contour is shown as a yellow dashed line. \textbf{(c)}~The myocardium is annotated in CAMUS but not EchoNet, so it is borrowed from CAMUS and transferred onto EchoNet; with no target ground truth, the transferred wall is shown against EchoNet's own LV-cavity contour (magenta) for anatomical context.}
\label{fig:cross_dataset}
\end{figure}

Because Match4Annotate operates on semantic feature representations, it can transfer annotations beyond the dataset in which they were originally defined, provided that the same anatomical structure is visible in the target video. Figure~\ref{fig:cross_dataset}
demonstrates this between EchoNet and CAMUS-TED. Since both datasets include left ventricular cavity annotations, we evaluate transfer in both directions, EchoNet$\rightarrow$CAMUS and CAMUS$\rightarrow$EchoNet, with ground truth available for quantitative comparison. In panels~(a) and~(b), the transferred cavity (green) follows the target ground-truth contour (yellow dashed) even though source and target are different subjects, imaged at different points of the cardiac cycle and with different sector geometry, depth, and zoom.
Table~\ref{tab:cross_dataset} quantifies the LV-cavity case. Match4Annotate transfers the cavity with a Dice score of $71.9$ on EchoNet$\rightarrow$CAMUS and $66.8$ on CAMUS$\rightarrow$EchoNet, the best one-shot result in both directions, ahead of Matcher (SemSAM) by $19.5$ and $16.6$ Dice. From a single source frame it also surpasses UniverSeg 3-shot in both directions and is within $1.1$ Dice of UniverSeg 5-shot on EchoNet$\rightarrow$CAMUS while exceeding it by $0.3$ on the reverse direction, trailing 10-shot by $4.6$ and $1.6$ Dice. Crossing datasets costs $6.6$ Dice on CAMUS-TED targets and $9.5$ on EchoNet targets relative to the within-dataset results of Table~\ref{tab:inter_dice}, though the two settings score slightly different frame sets.
We further transfer the myocardial mask from CAMUS-TED, where it is annotated, to EchoNet, where no myocardial ground truth is provided. Since no target ground truth exists, Figure~\ref{fig:cross_dataset}(c) shows the transferred wall against EchoNet's own LV-cavity contour (magenta), anatomically the inner boundary of the myocardium. The transferred band forms a closed ring that encloses this contour, with the largest deviations near the apex and the basal segments where the EchoNet sector truncates the wall. Such transfers are plausible labels rather than validated ones, but they illustrate the potential to reuse annotations across datasets with different labeling protocols.
\begin{table}[!t]
\centering
\caption{\textbf{Backbone ablation} on EchoNet cross-video transfer (200 pairs). The frozen feature extractor supervising $f_\theta$ is varied; all other components are fixed. Best in \textbf{bold}.}
\label{tab:backbone}
{\small\setlength{\tabcolsep}{1pt}
\begin{tabular}{lcccc}
\toprule
Backbone & \shortstack{PCK$\uparrow$\\.016/.031/.063} & Dice$\uparrow$ \\
\midrule
ImageNet ViT-S/16 & $1.8 / 6.6 / 20.6$ & $51.4{\pm}20.3$ \\
CLIP ViT-B/16 & $2.5 / 8.9 / 27.5$ & $61.0{\pm}17.6$ \\
DINOv2 ViT-S/14 & $2.9 / 10.3 / 30.8$ & $62.3{\pm}20.7$ \\
SAM~2 image encoder & $5.2 / 17.6 / 43.4$ & $69.3{\pm}15.6$ \\
DINOv3 ViT-S/16 (ours) & $\mathbf{6.1 / 18.4 / 46.7}$ & $\mathbf{76.3{\pm}12.3}$ \\
\bottomrule
\end{tabular}}
\end{table}
\begin{table}[!t]
\centering
\caption{\textbf{Ablation studies} on EchoNet cross-video transfer (200 pairs). \emph{Feature INR}: the spatiotemporal implicit feature field (Sec.~\ref{sec:feature}); \emph{Flow Prior}: the flow-guided spatial prior (Sec.~\ref{sec:flow} and ~\ref{sec:transfer}). \textbf{Bold} best, \underline{underline} second best; the full model is our configuration.}
\label{tab:ablation}
{\small\setlength{\tabcolsep}{1pt}
\begin{tabular}{lcccc}
\toprule
Variant & \shortstack{Feat.\\INR} & \shortstack{Flow\\Prior} & \shortstack{PCK$\uparrow$\\.016/.031/.063} & Dice$\uparrow$ \\
\midrule
Full model & \checkmark & \checkmark & $\mathbf{6.1}/\underline{18.4}/\mathbf{46.7}$ & $\mathbf{76.3{\pm}12.3}$ \\
\midrule
\multicolumn{5}{l}{\textit{Matching Strategy:}} \\
\quad w/o prior & \checkmark & & $5.2/16.2/39.9$ & $71.4{\pm}14.4$ \\
\quad prior from source & \checkmark & & $2.0/7.9/25.0$ & $63.4{\pm}14.8$ \\
\midrule
\multicolumn{5}{l}{\textit{Feature Representation:}} \\
\quad Low-Res ($28^2$) & & \checkmark & $3.4/12.6/36.8$ & $72.2{\pm}12.0$ \\
\quad Large Img ($112^2$) & & \checkmark & $4.6/14.9/40.3$ & $73.9{\pm}10.7$ \\
\quad FeatUp (DINOv3) & &  & $3.4/11.0/29.9$ & $61.8{\pm}17.0$ \\
\quad FeatUp (DINOv3) & & \checkmark & \underline{$5.9$}/$\mathbf{18.6}$/\underline{$45.0$} & $\underline{74.9{\pm}13.8}$ \\
\quad AnyUp (DINOv3) & &  & $4.2/14.2/36.9$ & $67.0{\pm}14.2$ \\
\quad AnyUp (DINOv3) & & \checkmark & $\underline{5.9}/17.3/43.5$ & $74.8{\pm}13.6$ \\
\bottomrule
\end{tabular}}
\end{table}
\subsection{Ablation Studies}
\label{sec:ablation}
\subsubsection{Backbone Ablation.}
\label{sec:backbone}
Table~\ref{tab:backbone} varies the frozen VFM supervising the feature field. DINOv3 is best on PCK and Dice across the board, supporting our backbone choice; the SAM~2 image encoder is the strongest alternative, while ImageNet \cite{dosovitskiy2021vit} and CLIP \cite{radford2021clip} encoders transfer less well to ultrasound.
\subsubsection{Method Ablation}
Table~\ref{tab:ablation} evaluates the two main design choices, the flow-guided prior and the feature representation, on EchoNet cross-video transfer (200 pairs).
\textit{Flow prior.} Removing the prior (``w/o prior'', \ie\ $\kappa \equiv 1$) and relying on feature matching alone reduces PCK@$.063$ from $46.7$ to $39.9$ and Dice from $76.3$ to $71.4$, showing that flow-guided spatial regularization improves correspondence. Anchoring the prior at the source position instead of the flow estimate (``prior from source'', $\hat{\mathbf{p}} = \mathbf{p}^s$) is worse still, at $25.0$ PCK@$.063$ and $63.4$ Dice, indicating that the gain comes from the learned, target-conditioned flow rather than from spatial locality alone. The prior is not specific to our feature field: adding it on top of FeatUp raises Dice from $61.8$ to $74.9$, and on top of AnyUp from $67.0$ to $74.8$. It is thus the largest single contributor in every representation we tested.
\textit{Feature representation.} Using DINOv3 features directly, either at the native low resolution ($28{\times}28$) or from large-image inference ($112{\times}112$), costs $2.4$--$4.1$ Dice and $6.4$--$9.9$ PCK@$.063$ relative to the continuous SIREN field, so the benefit is not merely one of feature-map resolution. Compared against learned upsamplers under the same flow prior, our field remains the strongest overall: it leads on Dice ($76.3$, vs.\ $74.9$ for FeatUp and $74.8$ for AnyUp) and on PCK@$.016$ and PCK@$.063$, with FeatUp marginally ahead only at PCK@$.031$ ($18.6$ vs.\ $18.4$).
\section{Conclusion}
\label{sec:conclusion}
We introduced Match4Annotate, which transfers point and mask annotations across independently acquired ultrasound videos by combining continuous spatiotemporal DINOv3 feature fields with feature-guided flow estimation, reaching state-of-the-art cross-video transfer on four cardiac and musculoskeletal datasets without task-specific offline training.
Transfer also extends across datasets with shared anatomy: between EchoNet-Dynamic and CAMUS-TED the method carries left-ventricular annotations in both directions, and carries myocardial labels from CAMUS-TED onto EchoNet, where they are unavailable, showing how annotations can be reused across labeling protocols.
The same formulation also propagates points and masks within individual ultrasound sequences. The Supplementary Materials show competitive performance against specialized tracking methods, despite Match4Annotate using no dedicated point- or mask-tracking architecture. This secondary capability suggests that a single annotation-transfer framework can support both cross-video transfer and within-video propagation.
\subsubsection{Limitations.}
Match4Annotate requires source and target videos to contain a shared anatomical structure that is captured by the pretrained features. Large out-of-plane motion, substantial acquisition differences, or partial visibility may therefore reduce correspondence quality. The smooth displacement field does not explicitly model occlusion or enforce topology preservation, and test-time optimization adds several minutes of computation per video. Moreover, propagated annotations in datasets without corresponding ground truth should be treated as plausible labels requiring further validation.
\subsubsection{Broader Impact.}
By enabling point and mask transfer across videos and datasets, Match4Annotate offers a practical path toward reusing scarce expert annotations across heterogeneous ultrasound collections.
\bibliographystyle{match4annotate-refs}
\bibliography{references}
\clearpage
\section*{Supplementary Materials}
\setcounter{figure}{0}
\setcounter{table}{0}
\setcounter{equation}{0}
\renewcommand{\thefigure}{S\arabic{figure}}
\renewcommand{\thetable}{S\arabic{table}}
\renewcommand{\theequation}{S\arabic{equation}}
\vspace{0.4em}
\noindent\textbf{Contents}
\vspace{0.15em}

{\small
\begin{description}[nosep,leftmargin=1.35em,labelwidth=1.2em,labelsep=0.15em,font=\normalfont\bfseries]
\item[A] \textbf{INR architecture.} Why SIREN, against ReLU, ReLU+PE and
  Instant-NGP, on feature-field reconstruction.
\item[B] \textbf{Networks and hyperparameters.} The feature field $f_\theta$,
  the downsampler $\mathcal{D}$, the flow field $g_\phi$, and every constant
  the main paper names.
\item[C] \textbf{Mask transfer details.} Interior-point extraction, KDE reconstruction,
  and the joint $(\sigma_\text{KDE},\tau)$ grid search.
\item[D] \textbf{Extended qualitative examples.} Cross-video point and mask transfer.
\item[E] \textbf{Within-video propagation.} Against specialized point
  trackers and video segmenters.
\item[F] \textbf{Time and memory.} Per-component and end-to-end cost.
\end{description}
}

\appendix

\section{Justification of SIREN over Alternative INR Architectures}
\label{sec:siren_justification}

We choose SIREN~\cite{sitzmann2020siren} over three alternative implicit neural representations: (1)~a ReLU MLP, (2)~a ReLU MLP with sinusoidal positional encoding (ReLU+PE)~\cite{mildenhall2020nerf}, and (3)~Instant-NGP with multi-resolution hash encoding~\cite{muller2022instant}. We compare the four on 5 randomly selected EchoNet-Dynamic videos under identical training budgets (500 epochs). For ReLU+PE we use $L{=}3$ positional-encoding frequencies (higher values introduce crosshatch artifacts). All models use 2 hidden layers of 256 dimensions; Instant-NGP additionally uses a 16-level multi-resolution hash grid.

\begin{table}[!htbp]
\caption{\textbf{Architecture comparison.} Reconstruction quality on 5 EchoNet-Dynamic videos under an identical 500-epoch budget (mean $\pm$ std). All four use 2 hidden layers of width 256; SIREN uses $\omega{=}30$, ReLU+PE uses $L{=}3$ positional-encoding frequencies~\cite{mildenhall2020nerf} (higher values introduce crosshatch artifacts), and Instant-NGP~\cite{muller2022instant} adds a 16-level hash grid, which is where its extra $16.8$M parameters go. Best in \textbf{bold}.}
\label{tab:arch_comparison}
\centering
\setlength{\tabcolsep}{4pt}
\small
\begin{tabular}{lccc}
\toprule
Architecture & Params & Recon.\ Loss$\downarrow$ & RMSE$\downarrow$ \\
\midrule
\textbf{SIREN} (ours) & 165,520 & $\mathbf{0.0038}{\pm}0.0004$ & $\mathbf{0.0613}$ \\
ReLU MLP & 165,520 & $0.0237{\pm}0.0026$ & $0.1540$ \\
ReLU+PE ($L{=}3$) & 170,128 & $0.0122{\pm}0.0019$ & $0.1103$ \\
Instant-NGP & 16,950,160 & $0.0253{\pm}0.0058$ & $0.1591$ \\
\bottomrule
\end{tabular}
\end{table}

\begin{figure}[!htbp]
\centering
{\small
\makebox[\linewidth]{%
  \makebox[0.2\linewidth]{DINOv3}%
  \makebox[0.2\linewidth]{SIREN}%
  \makebox[0.2\linewidth]{ReLU}%
  \makebox[0.2\linewidth]{ReLU+PE}%
  \makebox[0.2\linewidth]{NGP}}}
\\[1pt]
\includegraphics[width=\linewidth]{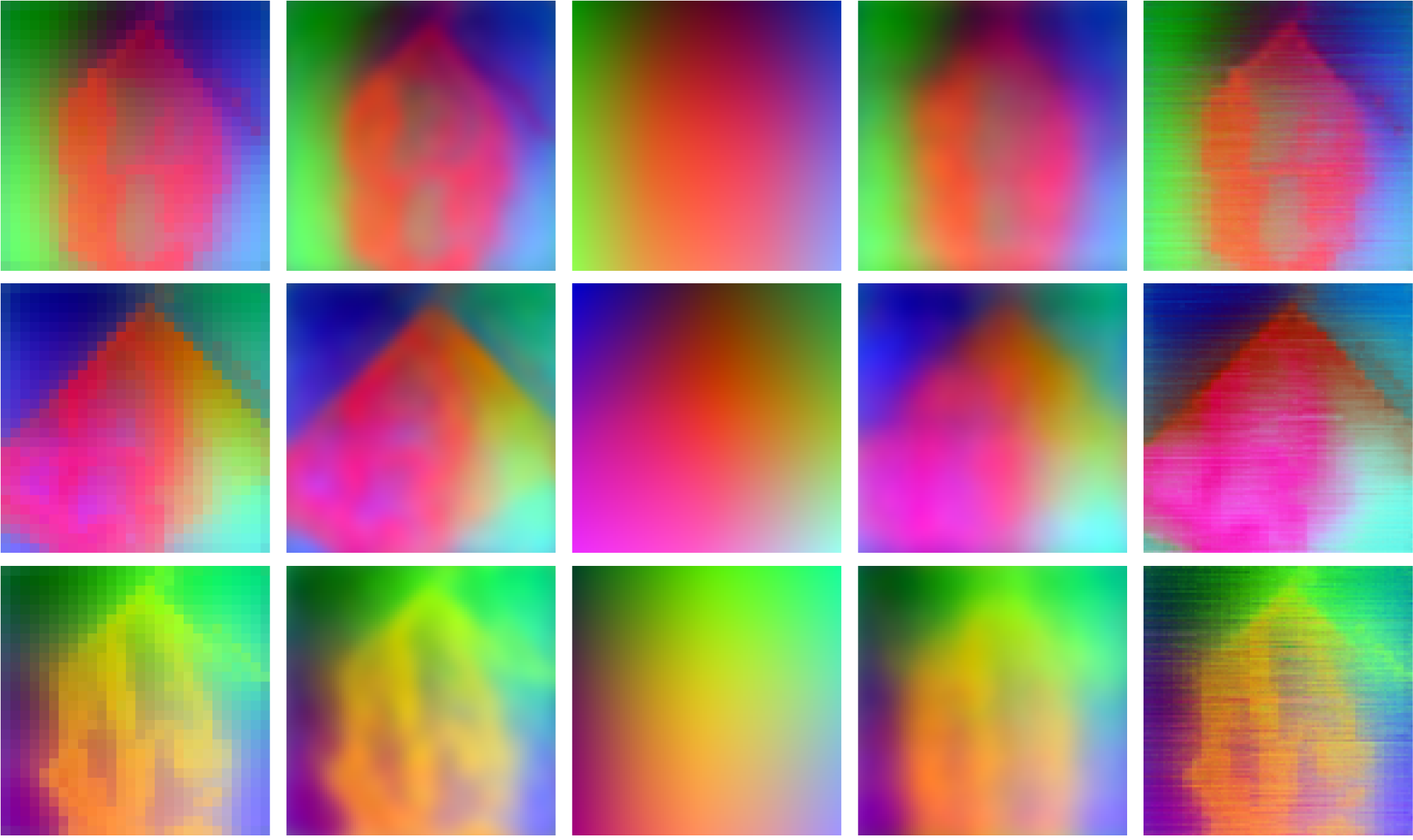}
\caption{\textbf{PCA feature maps.} RGB visualization of the first three PCA components of the reconstructed feature field, for the first frame of three EchoNet-Dynamic videos (one per row). Columns, left to right: the low-resolution DINOv3 target, then SIREN, ReLU MLP, ReLU+PE and Instant-NGP. SIREN produces the most faithful reconstruction; the alternatives show blocky artifacts (ReLU MLP, ReLU+PE) or fragmented features (Instant-NGP).}
\label{fig:pca_features}
\end{figure}

SIREN achieves $3{\times}$ lower reconstruction loss than the next-best alternative (ReLU+PE) with $100{\times}$ fewer parameters than Instant-NGP. ReLU+PE with reduced frequencies ($L{=}3$) improves over the default $L{=}10$ by mitigating crosshatch artifacts but still lags SIREN. Instant-NGP, despite ${\sim}17$M parameters, fails to learn smooth feature fields suitable for continuous coordinate queries. Figure~\ref{fig:pca_features} shows PCA-projected feature maps, where SIREN yields the most faithful, smooth reconstruction while ReLU variants show blocky artifacts and Instant-NGP fragmented features.

\subsubsection{Relation to the feature-representation ablation.}
Table~\ref{tab:arch_comparison} compares INR \emph{architectures} on how well
they reconstruct the feature field, measured by reconstruction loss and RMSE.
This is a different question from Table~\ref{tab:ablation}, which compares feature
\emph{representations} on the downstream transfer task, and there measures our
SIREN field against the FeatUp~\cite{fu2023featup} and
AnyUp~\cite{wimmer2025anyup} upsamplers under the same flow prior. The two are
complementary: this section motivates the choice of SIREN as the field
parameterization, while Table~\ref{tab:ablation} shows what that field is worth for
annotation transfer.

\FloatBarrier
\section{Network Architecture and Training Details}
\label{sec:architecture}

Match4Annotate uses three networks: a feature field SIREN $f_\theta$ (the
spatiotemporal implicit feature representation, Sec.~\ref{sec:feature}), a learned
convolutional downsampler $\mathcal{D}$ (used during training to bridge the
resolution gap between the INR output and the VFM supervision), and a
displacement SIREN $g_\phi$ (the continuous implicit feature flow,
Sec.~\ref{sec:flow}).

\begin{table}[!htbp]
\caption{\textbf{Complete hyperparameters} for the two implicit networks: the spatiotemporal feature field $f_\theta$ (Sec.~\ref{sec:feature}) and the continuous flow field $g_\phi$ (Sec.~\ref{sec:flow}). Both are optimized per video at test time with Adam~\cite{kingma2014adam} ($\beta_1{=}0.9$, $\beta_2{=}0.999$) at learning rate $10^{-4}$, with the backbone frozen. This table lists every quantity Sec.~\ref{sec:experimental_setup} names. $f_\theta$'s reconstruction grid is $112{\times}112$ for EchoNet-Dynamic, EchoNet-LVH and CAMUS-TED, and $224{\times}224$ for the higher-resolution MSK-Bone. $\lambda_1$ and $\lambda_2$ weight the smoothness (TV) and magnitude ($\ell_1$) terms of the flow loss; $d_\text{min}$ is the interior-point margin of Eq.~\eqref{eq:edt}; $\sigma_\kappa$ is the matching-prior bandwidth of Eq.~\eqref{eq:match}, and $\sigma_\text{KDE}$/$\tau$ are per dataset in Table~\ref{tab:kde_params}.}
\label{tab:all_hparams}
\centering
\setlength{\tabcolsep}{3pt}
\small
\begin{tabular}{llcc}
\toprule
 & Quantity & $f_\theta$ & $g_\phi$ \\
\midrule
$H_\text{dim}$ & Hidden width & 256 & 128 \\
$L$ & Hidden layers & 2 & 1 \\
$D$ & Output dimension & 384 & 2 \\
$\omega$ & Activation frequency & 30.0 & 30.0 \\
-- & Training epochs & 500 & 1000 \\
-- & Batch size (coords) & 1024 & -- \\
\midrule
\multicolumn{2}{l}{\textit{Backbone \& grids}} & \multicolumn{2}{c}{} \\
-- & DINOv3 input & \multicolumn{2}{c}{$448{\times}448 \rightarrow 28{\times}28$} \\
-- & Reconstruction grid & \multicolumn{2}{c}{$112^2$ / $224^2$ (MSK-Bone)} \\
\midrule
\multicolumn{2}{l}{\textit{Losses \& transfer}} & \multicolumn{2}{c}{} \\
$\lambda_1$ & Smoothness (TV) & \multicolumn{2}{c}{10.0} \\
$\lambda_2$ & Magnitude ($\ell_1$) & \multicolumn{2}{c}{0.01} \\
$d_\text{min}$ & Interior margin & \multicolumn{2}{c}{2\,px} \\
$\sigma_\kappa$ & Prior bandwidth & \multicolumn{2}{c}{$0.1\min(H,W)$} \\
\bottomrule
\end{tabular}
\end{table}

\subsection{Feature Field SIREN ($f_\theta$)}

\subsubsection{Architecture.}
$f_\theta : \mathbb{R}^3 \to \mathbb{R}^D$ is a SIREN with periodic activation $\rho(a) = \sin(\omega a)$, comprising $L$ fully-connected hidden layers of dimension $H_\text{dim}$ and a final linear projection to $D$. For input coordinate $\mathbf{c} = (x,y,t) \in [-1,1]^3$:
\begin{align}
\mathbf{h}_0 &= \sin(\omega (\mathbf{W}_0 \mathbf{c} + \mathbf{b}_0)), \notag \\
\mathbf{h}_l &= \sin(\omega (\mathbf{W}_l \mathbf{h}_{l-1} + \mathbf{b}_l)), \;\; l = 1, \ldots, L{-}1, \notag \\
f_\theta(\mathbf{c}) &= \mathbf{W}_L \mathbf{h}_{L-1} + \mathbf{b}_L. \notag
\end{align}

\subsubsection{Weight Initialization.}
Following~\cite{sitzmann2020siren}, $W_0 \sim \mathcal{U}(-1/d_\text{in}, +1/d_\text{in})$ with $d_\text{in}=3$, and $W_l \sim \mathcal{U}(-\sqrt{6/n}/\omega, +\sqrt{6/n}/\omega)$ where $n$ is the layer fan-in; biases are zero.

\subsubsection{Training.}
We train $f_\theta$ jointly with the downsampler $\mathcal{D}$ to reproduce frozen DINOv3-ViT-S/16 features at high resolution using the reconstruction loss (Eq.~\eqref{eq:recon}). The backbone is run at $448{\times}448$ input, giving $28{\times}28$ patch tokens of dimension $D{=}384$; the choice of backbone is ablated in Table~\ref{tab:backbone}. Hyperparameters are in Table~\ref{tab:all_hparams}.

\subsection{Convolutional Downsampler ($\mathcal{D}$)}
The feature SIREN outputs on a high-resolution grid $\Omega_\text{HR}$ of size $H{\times}W$, while VFM supervision is at the native patch resolution $H'{\times}W'$. The downsampler $\mathcal{D} : \mathbb{R}^{H\times W\times D} \to \mathbb{R}^{H'\times W'\times D}$ bridges this gap. We use a learned depthwise convolution: a single kernel $\mathbf{K}\in\mathbb{R}^{k_H\times k_W}$ shared across all $D$ channels. To act as a proper weighted average, weights are made non-negative ($|K_{ij}|$) and normalized ($\tilde{K}_{ij}=|K_{ij}|/\sum_{i,j}|K_{ij}|$), keeping downsampled features in the same range and preventing instability. Kernel size follows the resolution ratio: stride $=H/H'$ and kernel size $=\text{stride}+\mathbf{1}[H \bmod H' > 0]$ (\eg, $224{\to}28$ gives stride 8, kernel $8{\times}8$). A learned downsampler of this form is also used by FeatUp~\cite{fu2023featup}; the difference in our setting is the field it supervises, which takes coordinates only, without image intensity (Sec.~\ref{sec:feature}).

\subsection{Displacement SIREN ($g_\phi$)}
$g_\phi : \mathbb{R}^2 \to \mathbb{R}^2$ predicts per-coordinate displacements $(\Delta x,\Delta y)=g_\phi(x,y)$. We use a single-hidden-layer (``wide'') SIREN acting as a learned Fourier basis; a separate $g_\phi$ is trained per source--target frame pair, optimized with the flow loss (Eq.~\eqref{eq:flow}). At inference, the predicted center $\hat{\mathbf{p}}=\mathbf{p}^s+g_\phi(\mathbf{p}^s)$ seeds the Gaussian spatial prior for flow-guided annotation transfer (Eq.~\eqref{eq:match}). Hyperparameters are in Table~\ref{tab:all_hparams}.

\subsection{Matching Prior Bandwidth}
\label{sec:kappa}
The prior kernel $\kappa$ of Eq.~\eqref{eq:match} is an isotropic Gaussian centered
at $\hat{\mathbf{p}}$. Its bandwidth adapts to the annotation canvas so that the
prior covers a fixed \emph{fraction} of the field of view rather than a fixed
number of pixels: $\sigma_\kappa = 0.1\,\min(H,W)$, \ie\ $11.2$\,px on the
$112{\times}112$ EchoNet-Dynamic canvas and $49.6$\,px on the $656{\times}496$
MSK-Bone canvas. This keeps the strength of the spatial prior comparable across
datasets of very different native resolution, which is what lets a single rule
cover all four without per-dataset tuning.

\FloatBarrier
\section{Mask Transfer via Interior-Point KDE}
\label{sec:kde}

\subsubsection{Interior Point Extraction.}
Given a source mask $M^s$, we compute each interior pixel's Euclidean distance transform (EDT) to the nearest background pixel,
\begin{equation}
\text{EDT}(x,y;M^s) = \min_{(x',y')\notin M^s} \sqrt{(x-x')^2 + (y-y')^2}, \notag
\end{equation}
and select points sufficiently far from the boundary, $\mathcal{I}^s = \{(x,y) \mid \text{EDT}(x,y;M^s) \geq d_\text{min}\}$ with $d_\text{min}=2$ pixels. This excludes ambiguous boundary pixels while densely covering the region (typically hundreds to thousands of points).

\subsubsection{Propagation and Reconstruction.}
All interior points are independently propagated to the target via flow-guided annotation transfer (Sec.~\ref{sec:transfer}), yielding $\mathcal{I}^t$, then converted to a probability map by KDE with a Gaussian kernel,
\begin{equation}
P(x,y) = \frac{1}{Z}\sum_{\mathbf{p}^*_j \in \mathcal{I}^t} \exp\!\Big({-}\tfrac{(x-x^*_j)^2 + (y-y^*_j)^2}{2\sigma_\text{KDE}^2}\Big), \notag
\end{equation}
where $Z$ normalizes $P$ to $[0,1]$. In practice we accumulate unit impulses at each propagated point on the canvas, convolve with a 2D Gaussian, and normalize by the maximum. The binary mask is $\hat{M}^t(x,y) = \mathbf{1}[P(x,y) \geq \tau]$.

\subsubsection{Parameter Selection.}
KDE exposes the bandwidth $\sigma_\text{KDE}$ and threshold $\tau$. Small
$\sigma_\text{KDE}$ or high $\tau$ yields conservative masks that may fragment
when points are sparse or misaligned; large $\sigma_\text{KDE}$ or low $\tau$
yields smoother, more connected masks that may over-expand. The two interact, so
they are tuned \emph{jointly} by grid search rather than one at a time, on a
held-out subset disjoint from the evaluation transfers. For the experimental results reported in the main paper, we determine fixed parameter values using a small validation set with 10 videos. Table~\ref{tab:kde_params}
gives the setting adopted for each dataset.

\begin{table}[!htbp]
\caption{\textbf{KDE grid search.} Bandwidth $\sigma_\text{KDE}$ and threshold $\tau$ can be tuned jointly per dataset.}
\label{tab:kde_params}
\centering
\setlength{\tabcolsep}{3pt}
\small
\begin{tabular}{lcc}
\toprule
Setting & Canvas & Adopted \\
\midrule
EchoNet-Dynamic & $112{\times}112$ & $6.0\ /\ 0.25$ \\
MSK-Bone & $656{\times}496$ & $2.0\ /\ 0.01$ \\
CAMUS-TED & $448{\times}448$  & $3.5\ /\ 0.15$ \\
Cross-dataset & $448{\times}448$  & $4.0\ /\ 0.10$ \\
\bottomrule
\end{tabular}
\end{table}

\subsubsection{Why interior points rather than a boundary polygon.}
Reconstructing the mask from densely sampled interior points provides
redundancy: the kernel smooths individual point errors, so the mask degrades
gracefully when a subset of points are mismatched. Propagating boundary points
and closing them into a polygon has no such redundancy: a single badly
matched boundary point distorts the contour, and the polygon can
self-intersect when the ordering is not preserved.

\FloatBarrier
\section{Additional Cross-Video Transfer Examples}
\label{sec:inter_video_examples}

\begin{figure}[!htbp]
\centering
\includegraphics[width=\linewidth]{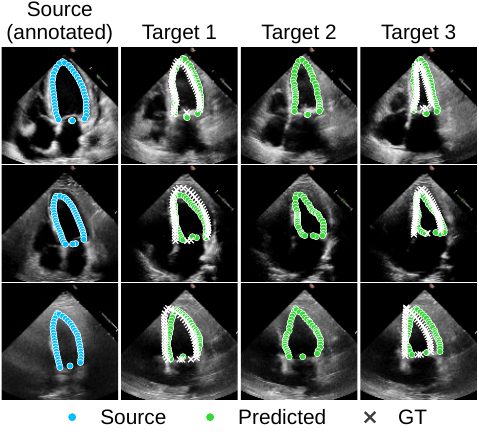}
\caption{\textbf{Cross-video point transfer, EchoNet-Dynamic.} Each row is one source$\rightarrow$target pair. Left: the source video's annotated frame. Right: its LV boundary points transferred onto three evenly spaced frames of a \emph{different} video. Ground truth is shown on the target frames that carry it (EchoNet-Dynamic labels only end-diastole and end-systole).}
\label{fig:inter_poi_echo}
\end{figure}

\begin{figure}[!htbp]
\centering
\includegraphics[width=\linewidth]{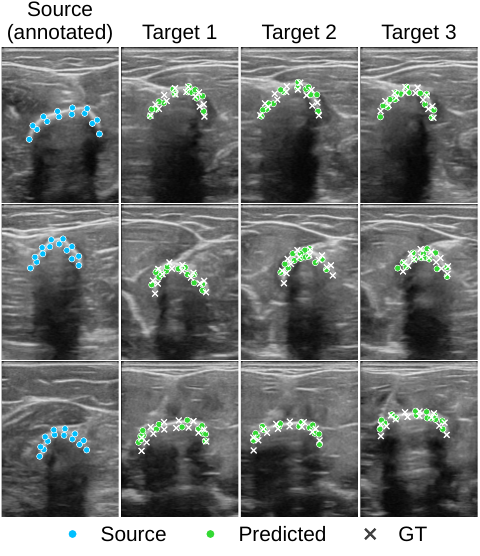}
\caption{\textbf{Cross-video point transfer, MSK-Bone.} As Figure~\ref{fig:inter_poi_echo}, for humerus boundary points transferred between subjects. The larger native canvas ($656{\times}496$) and the thin, elongated structure make this the harder geometry.}
\label{fig:inter_poi_msk}
\end{figure}

\begin{figure}[!htbp]
\centering
\includegraphics[width=\linewidth]{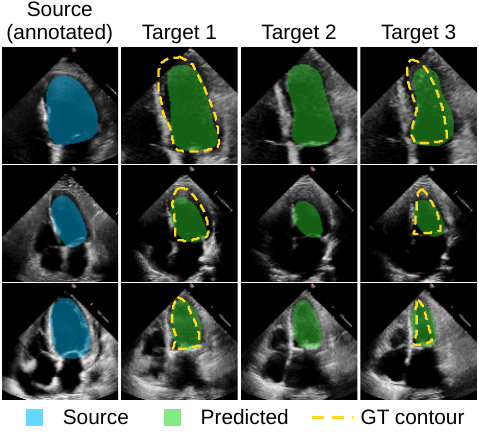}
\caption{\textbf{Cross-video mask transfer, EchoNet-Dynamic.} Left: the source video's annotated frame. Right: its LV-cavity mask transferred onto three evenly spaced frames of a \emph{different} video. Ground-truth contours appear on the target frames that carry them.}
\label{fig:inter_mask_echo}
\end{figure}

\begin{figure}[!htbp]
\centering
\includegraphics[width=\linewidth]{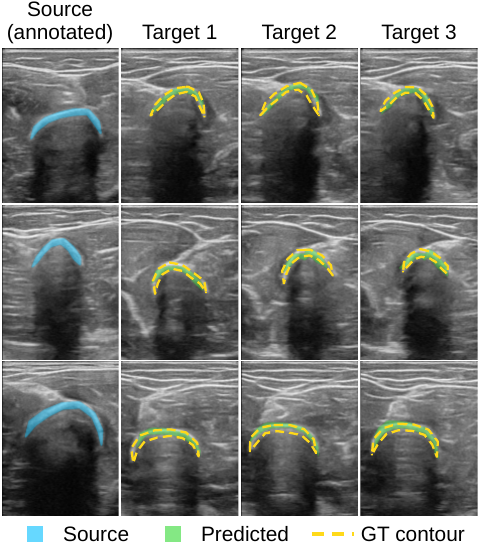}
\caption{\textbf{Cross-video mask transfer, MSK-Bone.} As Figure~\ref{fig:inter_mask_echo}, for humerus masks transferred between subjects. The thin, non-convex structure is the case the KDE reconstruction finds hardest, which is why MSK-Bone Dice trails the cardiac datasets in Table~\ref{tab:inter_dice}.}
\label{fig:inter_mask_msk}
\end{figure}

\begin{figure}[!htbp]
\centering
\includegraphics[width=\linewidth]{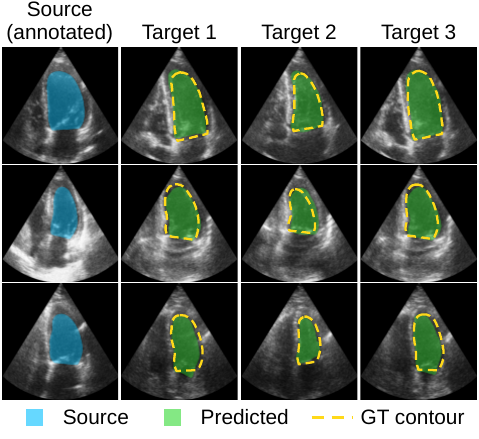}
\caption{\textbf{Cross-video mask transfer, CAMUS-TED.} Left: the source subject's annotated end-diastolic frame. Right: the LV-cavity mask transferred onto three evenly spaced frames of a \emph{different} subject's sequence, spanning its full cardiac cycle --- cross-phase transfer made visible.}
\label{fig:camus_masks}
\end{figure}

\begin{figure}[!htbp]
\centering
\includegraphics[width=\linewidth]{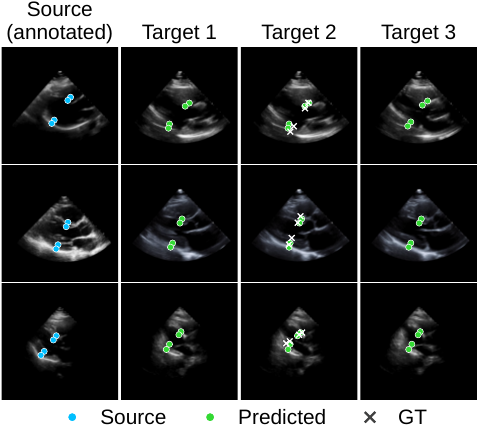}
\caption{\textbf{Cross-video point transfer, EchoNet-LVH.} Left: the source video's annotated frame with the four canonical landmarks. Right: those landmarks transferred onto three evenly spaced frames of a \emph{different} video. EchoNet-LVH labels one frame per video, so ground truth appears in at most one target column; the others show propagation to unlabeled frames.}
\label{fig:lvh_poi}
\end{figure}

Figures~\ref{fig:inter_poi_echo}--\ref{fig:lvh_poi} give qualitative examples
for all four datasets, extending Figures~\ref{fig:poi_comparison} and~\ref{fig:mask_comparison} and showing frames
beyond those with ground truth.

Two patterns are worth noting across them. The cardiac masks
(Figures~\ref{fig:inter_mask_echo} and~\ref{fig:camus_masks}) follow the
ventricle closely even where the target is at a different point in its cycle
than the source, whereas MSK-Bone (Figure~\ref{fig:inter_mask_msk}) shows the
thin-structure failure mode: a boundary that is a few pixels thick leaves
little room for the KDE field to be wrong in, which is the gap visible in
Table~\ref{tab:inter_dice}. For points, EchoNet-LVH (Figure~\ref{fig:lvh_poi}) labels
only one frame per video, so ground truth appears in at most one target column;
the remaining columns show that the four landmarks stay anatomically stable
across the sequence even where no label exists to score against.

\FloatBarrier
\section{Within-Video Propagation}
\label{sec:intra_supp}
 This is not the primary contribution of Match4Annotate: specialized trackers and segmenters are designed for this regime and remain stronger, and they still require one manual annotation \emph{per video}. Match4Annotate is nonetheless competitive while uniquely supporting cross-video transfer in the same framework. On EchoNet-Dynamic it underperforms dedicated trackers; on MSK-POI it surpasses TAPNext but trails CoTracker3 and Track-On2; on MSK-Bone the $\delta_\text{avg}$ gap narrows, approaching CoTracker3. For mask propagation it is close to SAM~2 on EchoNet-Dynamic but weaker on MSK-Bone, whose thin, non-convex structure is sensitive to the KDE step.

\begin{table}[!htbp]
\caption{\textbf{Within-video point propagation}, $\delta_\text{avg}\uparrow$ on EchoNet-Dynamic (100 videos), MSK-POI (36 videos) and MSK-Bone (20 videos). Baselines are the point trackers CoTracker3~\cite{karaev2024cotracker3}, TAPNext~\cite{tapnext} and Track-On2~\cite{trackon2}, none of which returns a mask. MSK-POI is the full 36-subject upper-arm set of~\cite{pallareslopez2026multimodal}, of which MSK-Bone is the 20-subject mask-annotated subset. Best in \textbf{bold}, second \underline{underlined}.}
\label{tab:intra_poi}
\centering
\setlength{\tabcolsep}{3pt}
\small
\begin{tabular}{lccc}
\toprule
Method & EchoNet & MSK-POI & MSK-Bone \\
\midrule
CoTracker3  & $\mathbf{41.8{\pm}10.6}$ & $\mathbf{54.7{\pm}7.4}$ & $\mathbf{40.9{\pm}7.4}$ \\
TAPNext     & \underline{$41.6{\pm}11.3$} & $27.3{\pm}7.5$ & $29.8{\pm}12.2$ \\
Track-On2   & $34.9{\pm}9.6$ & \underline{$47.0{\pm}8.3$} & $32.4{\pm}9.4$ \\
\midrule
Ours        & $31.7{\pm}11.5$ & $41.8{\pm}8.8$ & \underline{$38.2{\pm}7.0$} \\
\bottomrule
\end{tabular}
\end{table}

\begin{table}[!htbp]
\caption{\textbf{Within-video mask propagation}, Dice$\uparrow$ on the two datasets with masks. The baseline is the video object segmenter SAM~2~\cite{ravi2024sam2}, which returns no point correspondence. Best in \textbf{bold}.}
\label{tab:intra_mask}
\centering
\setlength{\tabcolsep}{3pt}
\small
\begin{tabular}{lcc}
\toprule
Method & EchoNet & MSK-Bone \\
\midrule
SAM~2 & $\mathbf{86.7{\pm}5.5}$ & $\mathbf{86.3{\pm}2.3}$ \\
\midrule
Ours  & $85.1{\pm}6.5$ & $74.2{\pm}5.7$ \\
\bottomrule
\end{tabular}
\end{table}

\begin{figure}[!htbp]
\centering
\includegraphics[width=\linewidth]{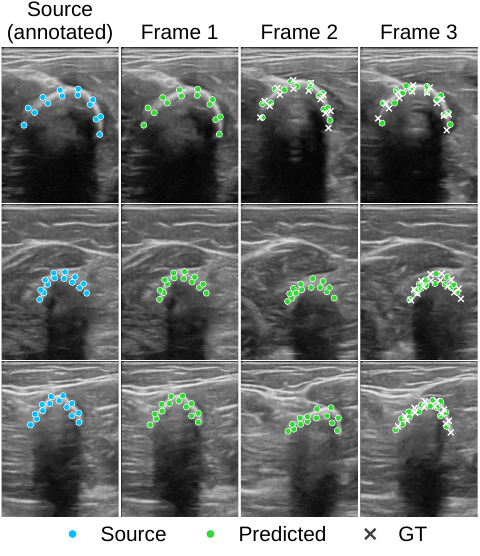}
\caption{\textbf{Within-video point propagation, MSK-Bone.} Left: the video's first labeled frame. Right: its points propagated to three evenly spaced frames of the \emph{same} video. Ground truth is shown on the frames that carry it.}
\label{fig:intra_poi}
\end{figure}

Figure~\ref{fig:intra_poi} shows temporal strips for within-video point tracking on MSK-POI, and Figure~\ref{fig:intra_masks} for mask propagation on EchoNet-Dynamic and MSK-Bone; predictions track landmarks smoothly across the video timeline from a single labeled source frame.

\begin{figure}[!htbp]
\centering
\includegraphics[width=\linewidth]{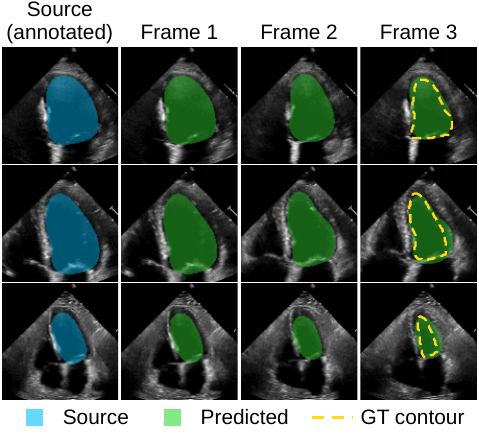}
\caption{\textbf{Within-video mask propagation, EchoNet-Dynamic.} Left: the video's first labeled frame. Right: its mask propagated to three evenly spaced frames of the \emph{same} video.}
\label{fig:intra_masks_echo}
\end{figure}

\begin{figure}[!htbp]
\centering
\includegraphics[width=\linewidth]{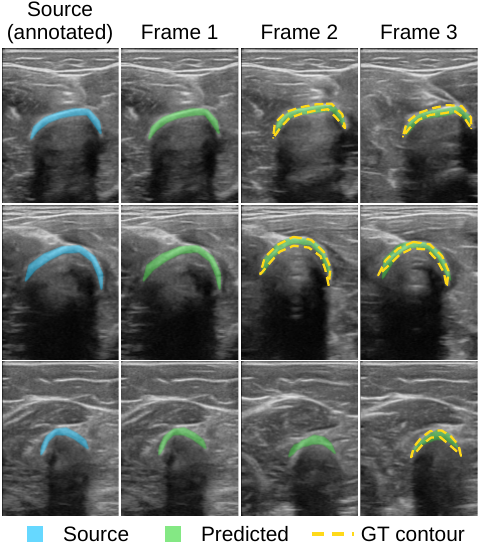}
\caption{\textbf{Within-video mask propagation, MSK-Bone.} As Figure~\ref{fig:intra_masks_echo}, within a single subject's sequence.}
\label{fig:intra_masks}
\end{figure}

\FloatBarrier
\section{Time and Memory Analysis}
\label{sec:timing}

Table~\ref{tab:timing} reports wall-clock time and peak GPU memory for each stage of the pipeline: feature extraction $\to$ feature SIREN training $\to$ flow SIREN training $\to$ flow-guided transfer $\to$ mask reconstruction.

\begin{table}[!htbp]
\caption{\textbf{Per-component timing and memory} on a single RTX~4090, mean $\pm$ std over 3 warmed-up repetitions on EchoNet-Dynamic. Measured over a 16-frame clip with DINOv3-ViT-S/16 at $448$\,px input and a $112{\times}112$ reconstruction grid; the feature SIREN is trained for 500 epochs and each flow SIREN for 1000 epochs on one source--target pair. Transfer costs are quoted for 42 boundary points (POI) and 500 interior points (mask).}
\label{tab:timing}
\centering
\setlength{\tabcolsep}{4pt}
\small
\begin{tabular}{lrr}
\toprule
Component & Time & VRAM (GB) \\
\midrule
Feature extraction   & $63{\pm}2$\,ms      & 0.23 \\
Feature SIREN        & $29.2{\pm}1.2$\,s   & 3.03 \\
Flow SIREN           & $0.8{\pm}0.01$\,s   & 0.17 \\
POI transfer         & $17{\pm}0$\,ms      & $<$0.1 \\
Mask transfer        & $158{\pm}4$\,ms     & $<$0.1 \\
KDE reconstruction   & $155{\pm}1$\,ms     & $<$0.1 \\
\bottomrule
\end{tabular}
\end{table}

Feature SIREN training dominates the per-video cost (${\sim}30$\,s). Each flow SIREN is trained \emph{once} per source--target pair and shared across all annotation points, so the $0.8$\,s flow cost is amortized whether propagating 42 contour points or 500 interior points. Peak memory (${\sim}3$\,GB, from feature SIREN training at $112{\times}112$) is well within mid-range GPUs; all other components use $<$0.25\,GB. For batch evaluation, a \texttt{torch.func.vmap}-based parallel training of $N$ independent 2D flow SIRENs gives ${\sim}14{\times}$ speedup over sequential training. Overall, the one-time ${\sim}30$\,s per-video cost is slower than feed-forward methods, but once trained each additional target frame costs only ${\sim}0.8$\,s---practical for annotating one frame and propagating to many.

\subsubsection{How this relates to the ``minutes per video'' figure.}
The timings in Table~\ref{tab:timing} and Table~\ref{tab:complexity} are measured
on EchoNet-Dynamic, the cheapest of the four datasets: a $112{\times}112$
reconstruction grid over a 16-frame clip. Cost grows with both the
reconstruction resolution and the number of frames the field must cover, and on
the larger datasets, i.e., MSK-Bone reconstructs at $224{\times}224$ over sequences
of up to several hundred frames, per-video fitting reaches several minutes.
The ``minutes per video'' figure quoted in the main paper refers to that
end-to-end, whole-sequence cost, whereas the ${\sim}59$\,s below is a single
source--target task on the smallest canvas.

\subsection{Runtime Comparison with Baselines}
\label{sec:complexity}

Table~\ref{tab:complexity} compares end-to-end runtime and memory against representative baselines on a single RTX~4090 (EchoNet-Dynamic, 5-run average). Match4Annotate performs test-time optimization: 59.4\,s for one cross-video task (two Feature SIRENs plus propagation), using $\sim$3\,GB of GPU memory. This is slower than feed-forward baselines because Feature SIREN training dominates ($\sim$30\,s per video); subsequent per-frame propagation costs only $\sim$0.8\,s, and adding more annotation points to the same pair is essentially free. The cost is one-time and amortized: a Feature SIREN trained once is reused across all source--target pairs for that video.

\begin{table}[!htbp]
\caption{\textbf{Runtime and memory} for one cross-video transfer task on a single RTX~4090 (EchoNet-Dynamic, 5-run average). Match4Annotate trades feed-forward speed for cross-video transfer; the dominant cost (Feature SIREN training) is amortized across targets.}
\label{tab:complexity}
\centering
\setlength{\tabcolsep}{4pt}
\small
\begin{tabular}{lcc}
\toprule
Method & Time$\downarrow$ & GPU Mem \\
\midrule
UniverSeg & 0.02\,s & 0.07\,GB \\
MATCHA & 24.3\,s & 12.3\,GB \\
Match4Annotate & 59.4\,s & $\sim$3\,GB \\
\bottomrule
\end{tabular}
\end{table}

\end{document}